\documentclass{article}

\usepackage{arxiv}
\pdfoutput=1
\usepackage{graphicx}
\usepackage{color}
\usepackage{subfigure}
\usepackage{tikz}
\usetikzlibrary{automata,calc,backgrounds,arrows,positioning, shapes.misc,quotes,arrows.meta, decorations.pathreplacing}
\usepackage[normalem]{ulem}
\usepackage{braket}
\usepackage{verbatim}
\usepackage{amssymb}
\usepackage{mathtools}
\usepackage{bm}
\usepackage{bbold}

\usepackage[utf8]{inputenc} 
\usepackage{csquotes}
\usepackage[T1]{fontenc}
\usepackage[english]{babel}
\usepackage{hyperref}       
\usepackage{url}           
\usepackage{booktabs}      
\usepackage{amsfonts, amsmath, amssymb}      
\usepackage{nicefrac}       
\usepackage{microtype}     
\usepackage{lipsum}
\usepackage[sorting=none, style=science, citestyle=phys]{biblatex}
\usepackage[affil-it]{authblk}
\newcommand{\beq}{\begin{equation}}
\newcommand{\eeq}{\end{equation}}
\newcommand{\bea}{\begin{eqnarray}}
\newcommand{\eea}{\end{eqnarray}}
\newcommand{\nn}{\nonumber \\}
\newcommand{\E}{\mathbb{E}}

\newcommand{\z}{\mathcal{\bzeta}}
\def\pd{\partial} 

\def\x{\mathbf{x}}

\def\s{\mathbf{s}}
\def\z{\mathbf{z}}
\def\u{\mathbf{u}}

\title{RBM-Flow and D-Flow: Invertible Flows with Discrete Energy Base Spaces}

\author[1, 2, 3, 4]{Daniel O'Connor}
\author[1, 2]{Walter Vinci}
\affil[1]{Quantum Artificial Intelligence Laboratory (QuAIL), NASA Ames Research Centre, Moffet Field, CA, USA}
\affil[2]{KBR, Jefferson St., Houston, TX, USA}
\affil[3]{Department of Electronic and Electrical Engineering, University College London, Gower Street, London, UK}
\affil[4]{London Centre for Nanotechnology, Gordon Street,
London, UK}
\date{}

\begin{document}
    \maketitle
    
    \begin{abstract}
    Efficient sampling of complex data distributions can be achieved using trained invertible flows (IF), where the model distribution is generated by pushing a simple base distribution through multiple non-linear bijective transformations. However, the iterative nature of the transformations in IFs can limit the approximation to the target distribution. In this paper we seek to mitigate this by implementing RBM-Flow, an IF model whose base distribution is a Restricted Boltzmann Machine (RBM) with a continuous smoothing applied. We show that by using RBM-Flow we are able to improve the quality of samples generated, quantified by the Inception Scores (IS) and Frechet Inception Distance (FID), over baseline models with the same IF transformations, but with less expressive base distributions. Furthermore, we also obtain D-Flow, an IF model with uncorrelated discrete latent variables. We show that D-Flow achieves similar likelihoods and FID/IS scores to those of a typical IF with Gaussian base variables, but with the additional benefit that global features are meaningfully encoded as discrete labels in the latent space.
    \end{abstract}
    
    \section{Introduction}
\label{sec:introduction}

Using generative models to efficiently and accurately represent high dimensional data has been widely explored with generative adversarial networks (GANs) \cite{goodfellow_generative_2014}, variational auto-encoders (VAEs) \cite{kingma_auto-encoding_2014}, energy-based models (EBM) \cite{du_implicit_2019, lecun2006tutorial}, diffusion models \cite{sohl-dickstein_deep_2015, ho_denoising_2020}, autoregressive models \cite{pmlr-v15-larochelle11a, pmlr-v48-oord16, chen_pixelsnail_2017, parmar_image_2018}, non-autoregressive invertible flows \cite{dinh_nice_2015, dinh_density_2017, kingma_glow_2018, ho_flow_2019}, and continuous normalizing flows \cite{grathwohl_ffjord_2018, chen_neural_2019}. Each approach presents unique advantages and challenges, depending on the type of dataset or application.

Non-autoregressive invertible flows (IFs) are a promising class of models that allow for exact likelihood training (unlike VAEs), can be efficiently sampled from (unlike autoregressive models), and do not require a discriminator network (unlike GANs). It does this by implementing a model that can tractably produce invertible, non-linear transformations as well as calculating its respective Jacobian by using partitioned bijective functions. However, they have so far failed to reach state-of-the-art density estimation benchmarks set by some of these other models. This issue can be attributed to the IF procedure resembling sequential local refinements rather than a global adjustment of the model distribution (see a visual demonstration of this in Ref.~\cite{grathwohl_ffjord_2018}).

Instead of looking to improve the expressiveness of our bijective transformations in IF models, in this paper we look at improving the expressiveness of the base distribution of our model. Typically, a Gaussian distribution of zero mean and unit variance is used as the base distribution in IF models, but we draw from the strengths of established EBMs into a new framework, which we call EBM-Flow. Our model looks to use an EBM prior to model global features, and then use an IF model for sequential refinement.

Joint adversarial training of IFs and EBMs was previously explored in Ref.~\cite{gao_flow_2020} in order to utilise the fact that both EBMs and IFs can be trained via maximum-likelihood estimation, and that EBMs do not make any assumption on the target distribution, which is modeled by a global scalar function using a relatively small number of parameters. However, the training of continuous EBMs requires expensive Markov Chain Monte Carlo (MCMC) techniques such as Hamiltonian Monte Carlo or Langevin dynamics~\cite{girolami_riemann_2011, song_chun_zhu_grade_1998}.
 
Discrete EBMs such as restricted Boltzmann machines (RBMs)~\cite{ackley1985learning} can be trained more efficiently by taking advantage of techniques such as persistent contrasting divergence (PCD) with block Gibbs sampling to allow for faster and more efficient MCMC mixing~\cite{tieleman_training_2008}. Therefore, they are less computationally expensive to train and scale compared to continuous EBMs. Furthermore, discrete EBMs can possibly take advantage of the emergence of new quantum technologies that can simulate and speed-up training of BMs with, for example, quantum annealers~\cite{vinci_path_2019}.

When integrating discrete EBMs with IF models, one could use IFs that operate with discrete variables~\cite{hoogeboom_integer_2019, tran_discrete_2019}, but their density estimation benchmarks are notably worse than IF models that operate in the continuous domain. Therefore, in our specific implementation, we look to perform a continuous smoothing of a discrete EBM as the base energy distribution, followed by a custom IF model (similar to the FLOW++~\cite{ho_flow_2019} architecture, and outlined in Appendix~\ref{sec:Appendix A: RBM-IF Architecture}) for sequential refinement. Specifically, we look at using an RBM as our discrete EBM due to the efficient mixing we can get from using PCD with block Gibbs sampling. There have been several attempts in the literature to smooth RBMs to model continuous variables~\cite{khoshaman_gumbolt_2018, chu_restricted_2018}. We found the most effective technique to be the Gaussian smoothing introduced early on in Ref.~\cite{zhang_continuous_2012}, and more recently in Ref.~\cite{hartnett_probability_2020}. The advantage of this technique is that it allows for unbiased, log-likelihood maximization of the smoothed RBM, since inference from the continuous to the discrete latent variables is exact. Other properties of the model are also retained such as efficient sampling and density estimation. This introduces the main model tested in this paper - RBM-Flow. 

To study the performance of RBM-Flow and the advantages of utilizing the smoothed RBM as base distribution for the flow, we perform several ablation experiments in which we keep the same structure for the IF but replace the smoothed RBM with less expressive base distributions. Among all the models considered, RBM-Flow is unique in the sense that it is the only model whose base distribution is capable of modeling multi-modal distributions. We thus show, both visually and quantitatively by evaluating the Inception Scores (IS) and Frechet Inception Distance (FID), that RBM-Flow is able to generate samples of better quality than other baseline models when trained on CIFAR10. Moreover, RBM-Flow is able to generate samples of higher quality with significantly lower training time.

We also focus on a particularly interesting ablation of the RBM-Flow model, D-Flow, which is obtained by simply setting all the couplings of the base RBM to zero. It turns out that the corresponding continuous base model of D-Flow is a product of a mixture of Gaussians with centers given by the values of a set of discrete latent variables $\s=\pm1$. D-Flow can thus be seen as an implementation of an IF with discrete latent variables, thus generalizing previous work done within the VAE framework~\cite{khoshaman_gumbolt_2018, vahdat_dvae_2018, sadeghi_pixelvae_2019, khoshaman_quantum_2018, rolfe_discrete_2017}. As far as we are aware, all the approaches involving VAEs require the implementation of smoothing techniques that result in training with respect to a biased loss function which is not guaranteed to be a lower bound to the likelihood~\cite{jang_categorical_2017, maddison_concrete_2017}. The advantage of the D-Flow implementation is that training is performed by maximization of the exact likelihood of a continuous model, which is connected to a discrete latent model via exact inference. As we show, D-Flow is also able to encode global features when trained on the CIFAR10 dataset, while achieving a log-likelihood similar to that of a standard IF with Gaussian base model. The use of generative models with discrete latent variables is appealing for various applications, including scenarios in which unsupervised or semi-supervised learning is used to encode meaningful global features of the dataset into a set of discrete labels.

In this paper, we outline the background to RBM-Flow in section~\ref{sec:background}, before then defining EBM-Flows and its construction in section~\ref{sec:eflow}. Experimental analysis is conducted using ablative tests in section~\ref{sec:experiments}, where we compare the RBM, Bernoulli, independent Gaussian, and Multivariate Gaussian base distributions. Finally we conclude in section~\ref{sec:conclusions}. 

    \section{Background}
\label{sec:background}

\subsection{Invertible Flow Models}
\label{subsec:IF modelss}

The foundation of non-autoregressive invertible flow models (IF) is based on learning a differentiable, invertible, non-linear, bijective function (i.e. the flow) $f : \mathcal{X} \to \mathcal{Z}$, such that one can approximately map a base probability distribution $p_Z(\z)$ to the model distribution, $p_X(\x)$, with $\z = f(\x)$, using a change of variables:
\beq
\log p_X(\x) = \log p_Z(\z) + \log \left|\det \left( \frac{d\z}{d\x} \right) \right| = \log p_Z(f(\x)) + \log \left|\det \left( \frac{df}{d\x} \right) \right|.
\label{eq: change_of_variables}
\eeq

The usual choice is to have the base $\z$ variables independently distributed according to a product of Gaussian distributions with zero mean and unit variance. IF models can then be trained by maximizing the log-likelihood of data samples passed through the trainable flow, $f$, to the Gaussian prior. The key challenge is to provide an explicit parameterization of the flow that is powerful and expressive, and that also allows for efficient computation of the determinant of the Jacobian of $f$. These two properties are necessary to enable IFs to accurately model complex data distributions, and to be computationally scalable to high-dimensional datasets.

The first step towards building a flow $f$ with the required properties is to add depth to the model though a chain of $K$ simpler flows $f^k$ called affine coupling layers:
\bea
 &f  = f^K \circ f^{K-1} \dots \circ f^{2} \circ f^1\,;& \nn
 & \x^k \equiv \{\x^k_1, \x^k_2\}, \quad \x^0 \equiv \x, \quad \x^K \equiv \z, \quad \x^k \equiv f^k(\x^{k-1}) \,; &\nn
 & \{\x^k_1, \x^k_2\} = \{\x^{k-1}_1, \x^{k-1}_2 e^{a(\x^{k-1}_1)} + b(\x^{k-1}_1)\}\,, \quad \log \left|\det \left( \frac{d f^k}{d\x^{k-1}} \right) \right| = a(\x^{k-1}_1)\,. &
 \label{eq: affine_flow}
\eea
Notice that the affine coupling layer is the identity transformation over one subset of the partitioned variables, such that it can be used to condition the transformation of the other subset, and therefore retain a tractable Jacobian. This requires the flow to have a minimum of 4 coupling layers interleaved with permutations and/or 1x1 invertible convolutions~\cite{kingma_glow_2018} in order to fully transform a distribution. In general, recent state-of-the-art flows will also include invertible normalization layers and implement a multi-scale architecture via appropriate squeeze and reshape transformations~\cite{kingma_glow_2018, dinh_density_2017, ho_flow_2019}. 

There are two more key technical advancements that are worth discussing with respect to IF. The first one is the implementation of variational dequantization. Dequantization is an important procedure when training continuous probability models with discrete datasets, such as natural images stored with 8 or 16 bit precision. In fact, the log-likelihood of a continuous model on discrete data is unbounded, thus causing instabilities and divergences during training~\cite{uria_rnade_2013}. The standard technique to train on a bounded loss function is to add uniform random noise to the training data $\x \rightarrow \x + \u$. Variational dequantization improves this technique by using a trainable noise function as follows $\u \sim p_n(\u|\x)$. The corresponding loss function can be seen as a (tighter) variational lower bound to the log-likelihood of the model on the original dataset. As shown in Ref.~\cite{ho_flow_2019} this greatly improves regularization and generalization of the model.

The second improvement is the use of more expressive coupling layers. Much of the previous work has exploited simple yet effective affine transformations (Eq. \ref{eq: affine_flow})~\cite{dinh_nice_2015, dinh_density_2017, kingma_glow_2018}, but the state of the art held by Flow++ \cite{ho_flow_2019} uses parameterized cumulative distribution functions for a mixture of $K$ logistics (MixLogCDF):
\bea
    \label{eq: MLCDF}
    &\textrm{MixLogCDF}\left( x; \mathbf{\pi}, \mathbf{\mu}, \mathbf{s} \right) := \sum^K_{i=1} \pi_i \sigma \left( (x - \mu_i ) \cdot \exp(-s_i) \right) \, &\nn
& \{\x^k_1, \x^k_2\} = \{\x^{k-1}_1, \sigma^{-1}\left(\textrm{MixLogCDF}(\x^{k-1}_2; \mathbf{\pi}(\x^{k-1}_1), \mathbf{\mu}(\z_1), \mathbf{s}(\x^{k-1}_1))\right) \cdot \exp(a(\x^{k-1}_1)) + b(\x^{k-1}_1)\}\,.&
\eea

With these advancements IFs have been recently scaled to generate high-resolution natural images that compete with Generative Adversarial Networks (GAN) in visual quality. However, despite providing fast generation as with GANs and unlike autoregressive models, IFs require a large number of parameters and training is computationally expensive. 

We conclude this introduction to IFs by pointing out that IFs are typically discussed in the context of generative models with latent variable. IFs should be more appropriately considered as fully visible models because the relationship between $\x$ and $\z$ is deterministic, not probabilistic, and inference is not necessary.


\subsection{Energy Based Models}
\label{subsec:EBMs_background}

Energy based models (EBM) are defined by a scalar energy function, $E(\x)$, which represents a non-normalized log-probability for the configuration $\x$:
\beq
p(\x) = \frac{e^{-E(\x)}}{Z}\,, \quad Z = \int e^{-E(\x)} d\x\,.
\label{eq:EBM}
\eeq
The key observation for training energy models is that evaluation of the normalization constant (partition function) $Z$ is not necessary. Gradients can be computed as follows:
\beq
\pd \log p(\x) =  - \pd E(\x) - \pd \log Z = - \pd E(\x) + \E_{\x' \sim p(\x')}[ \pd E(\x')]\,.
\eeq
where the second term (the negative phase) is the expectation of the gradient of the energy function evaluated on the model samples $\x'$. However, when modelling high-dimensional distributions, the negative phase becomes more difficult to evaluate as it can develop sharp modes which makes evaluation computationally expensive due to the long mixing times in the MCMC methods employed.

Nonetheless, EBMs are very appealing for their mathematical elegance and ability to model complex probability distributions with global features using a relatively small number of parameters. Modern implementations of EBMs~\cite{du_implicit_2019, gao_flow_2020, pmlr-v100-du20a} have achieved impressive results on large scale datasets by using carefully designed deep neural networks to regularize the energy function and improve mixing of MCMC simulations, but heavily use expensive techniques such as Hamiltonian Monte Carlo and Langevin dynamics~\cite{girolami_riemann_2011, song_chun_zhu_grade_1998}.

\subsection{Boltzmann Machines}
\label{subsec:Boltzmann Machines}

The use of discrete EBMs are appealing due to their likeness to physical systems in nature. Boltzmann Machines (BM) are a class of discrete EBMs in which the space of configurations is now composed of an ensemble of spins $\s \in \{-1, 1\}^N$, and a quadratic energy functional:
\beq
p(\s) = \frac{e^{-E(\s)}}{Z}\, \quad E(\s) = - \frac12 \s^T J \s - h^T \s, \quad  Z = \sum_{\{\s\}}e^{-E(\s)} \,.
\eeq

The training of large BMs is still computationally expensive but it is significantly more efficient for a limiting case called the restricted Boltzmann machine (RBM). This is where the BM is restricted to a bipartite form composed of hidden units that support visible units from which we take samples from. With RBMs, we can scale to larger models due to the availability of less computationally expensive MCMC techniques such as contrastive divergence (CD)~\cite{hinton_fast_2006}, and Persistent CD (PCD)~\cite{tieleman_training_2008}, coupled with block Gibbs sampling.

\subsection{Gaussian Smoothing of Boltzmann Machines}
\label{subsec:Gaussian Smoothing}

Probability distributions of binary discrete variables can be efficiently modelled with BMs, therefore, there has been a long-time interest in the research community to adapt RBMs to model continuous variables too. The simplest approach trains a BM by interpreting continuous data as the expectations needed to compute the gradients of the RBM. Other more mathematically justified approaches involve fully continuous models such as the harmonium~\cite{smolensky_information_1986, welling_exponential_2005}, and continuous-discrete models such as RBMs with Gaussian visible units and discrete latent units~\cite{chu_restricted_2018}.

We are interested in a continuous smoothing of a BM which retains all the advantages in fast and efficient training with PCD sampling, and has a mathematically well-defined continuous formulation that enables log-likelihood training with continuous data. The smoothing technique we found ideal for our goals was introduced in Ref.~\cite{zhang_continuous_2012}, and it was used more recently in connection with a study on spin glasses in Ref.~\cite{hartnett_probability_2020}. This smoothing technique we describe now is a simple application of Gaussian square completion. Given a BM with the following energy function:
\beq
E(\s) = - \frac12 \s^T J \s - h^T \s\,,
\eeq
one introduces a set of continuous variables $\z$ such that:
\beq
\z \sim p(\z|\s) = \mathcal N(\mu = \s, \Sigma=\tilde J^{-1}), \quad \tilde J = J + \Delta \mathbb{1}\,,
\label{eq: discrete_smoothing}
\eeq
where $ \mathcal N$ is a multivariate Normal distribution with centers $\s$ and covariance matrix $\Sigma$. Hence we refer to the $\z$ variables as a Gaussian smoothing of the discrete $\s$ variables. The constant $\Delta$ must be chosen large enough to ensure that the covariance matrix $\Sigma$ is positive-definite. The log-probability of the joint distribution $p(\z, \s) = p(\z|\s) p(\s)$ can then be written as follows:
\bea
\log p(\z, \s) = -\frac12 \z^T \tilde J\z  + \s^T \tilde J \x + h^T\s - \log Z_{\z,\s}\,. 
\eea
The discrete variables $\s$ appear linearly in the joint distribution, such that it can be marginalized out to get:
\beq
\log p(\z) = -\frac12 \z^T \tilde J\z + \log \left[ 2 \cosh \left(\sum_i \tilde h_i(\z)\right)\right]- \log Z_{\z} \,, \quad \tilde h(\z) = \tilde J \z + h\,.
\eeq
We thus have smoothed the BM with the effective continuous scalar energy function defined above. Notice that sampling from the continuous model can be performed via ancestral sampling after sampling $\s$ first from the discrete BM:  $\z \sim p(\z|\s)$. Moreover, the normalization constant $\log Z_{\z}$ can be computed from the knowledge of the normalization constant of the BM, $\log Z_{\s}$, via the following relationship~\cite{hartnett_probability_2020}:
\beq
\log Z_{\z} =\log Z_{\s} + \frac{N}{2} \log 2\pi - \frac12\log \det{\tilde J} + \frac{N}{2} \Delta\,.
\eeq
Sampling and $\log Z$ evaluation of the discrete latent model is computationally less challenging, but has the trade-off of having to perform inversion of the $\tilde J$ matrix and compute the  $\log \det{\tilde J}$ respectively.
    \section{EBM-Flow}
\label{sec:eflow}

\subsection{EBM-Flow Formalism}
\label{subsec:eflow}

In this section we define the EBM-Flow, a class of EBMs whose scalar energy functional is obtained by a base energy functional via an IF-induced change of variables. The interpretation of EBM-Flow is realised through the generated samples of the target distribution still being samples from an EBM. It follows from Eqs.~\ref{eq: change_of_variables} and~\ref{eq:EBM} by definition:
\beq
\log p(\x) = - E_{\rm EBM}(f(\x)) - \log Z_{\rm EBM} + \log |\det (df/d\x) | \equiv - E_{\rm EBM-Flow}(\x) - \log Z_{\rm EBM-Flow}\,,
\eeq
where we have defined the EBM-Flow energy functional to be:
\beq
E_{\rm EBM-Flow}(\x) = E_{\rm EBM}(f(\x)) - \log |\det (df/d\x) |\,.
\eeq
The partition functions are also preserved through the change of variables:
\beq
Z_{\rm EBM} = \int e^{- E_{\rm EBM}(\z)} d\z = \int e^{- E_{\rm EBM} (f(\x))} \det (df/d\x) d\x = \int e^{- E_{\rm EBM-Flow}(\x)} d\x = Z_{\rm EBM-Flow}\,.
\eeq

Therefore one can see that by joining the EBM framework with an IF model, the IF transformations simply become an extension to the EBM, thus resulting in a model that utilises the strengths of both EBMs and IFs. This permits EBM-Flow to have a more complex energy function, $E_{\rm EBM-Flow}(\x)$, despite only sampling from a base EBM that corresponds to a potentially simpler base scalar energy functional, $E_{\rm EBM}(\z)$.

\subsection{RBM-Flow}
\label{subsec:rbm-flow}

RBM-Flow is a special implementation of an E-Flow in which the base model is a Gaussian smoothed RBM, as described in section~\ref{subsec:Gaussian Smoothing}:
\beq
E_{\rm RBM-Flow}(\x) \equiv  \frac12 f(\x)^T \tilde J f(\x) - \log \left[ 2 \cosh \left(\sum_i \tilde h_i(f(\x))\right)\right]  - \log |\det (df/d\x) |\,,
\eeq
where $\tilde J = J + \Delta \mathbb{1}$ is a positive-definite matrix, $J$ is chosen to be the connectivity of a Restricted Boltzmann machine, and $f$ is an IF of the Flow++ type (see details in the Appendix~\ref{sec:Appendix A: RBM-IF Architecture}, and in Ref.~\cite{ho_flow_2019}). Note that we cannot definitively know a priori what choice of $\Delta$ to use that ensures $\tilde J$ to be positive semi-definite during training, a condition required for the Gaussian smoothing.


The advantage of implementing RBM-Flow is that sampling from the continuous base model can be performed via ancestral sampling of the discrete $\s$ variables. This can be done very efficiently via block-Gibbs sampling and PCD. To obtain the continuous base variables $\z$, one has to sample from the multivariate Normal distribution $\z \sim  \mathcal N(\mu = \s, \Sigma=\tilde J^{-1})$. The cost of this operation is given by an inversion followed by a Cholesky decomposition of the matrix $\tilde J$, both of which have a complexity of $\mathcal O(N^3)$. Given that CIFAR10 only generated matrices with dimensions of a few thousand, these operations were carried out relatively fast. 

If scaling to large sizes is needed, one can remove the bipartite restriction of an RBM to either enable a fully connected BM, or an RBM where the visible units have inter-connectivity. This allows us to directly use a Cholesky-decomposed ansatz for the connectivity matrix: $\tilde J = CC^T$, at the cost of sampling from a fully connected BM instead. Moreover, an ansatz that enables efficient inversion of $\tilde J$ can be described by:
\beq
\tilde J = C C^T, \quad C = \left(\begin{matrix}
\mathbb{1} & W\\
0 & \mathbb{1}
\end{matrix} \right), \quad C^{-1} = \left(\begin{matrix}
\mathbb{1} & - W\\
0 & \mathbb{1} \end{matrix}  \right)\,,
\eeq
which describes an RBM where the visible layer now has inter-connectivity between units.

\subsection{D-Flow}

D-Flow is obtained by setting the coupling matrix of RBM-Flow to zero:
\beq
E_{\rm D-Flow}(\x) \equiv  \frac\Delta2 f(\x)^T f(\x) - \log \left[ 2 \cosh \left(\Delta \sum_i  f_i(\x) + h_i\right)\right] - \log |\det (df/d\x) |\,,
\eeq
where now $\Delta$ is a positive constant. Notice that the resulting continuous base model of D-Flow is a mixture of Gaussians $\z|\s \sim  \mathcal N(\mu = \s, \Sigma=\Delta^{-1})$, with the locations given by the discrete variables $\s$, which are distributed according to a set of independent Bernoulli variables with probabilities $p(\s=-1) = \rm{Sigmoid}(h)$. 

As we shall see in the next section, when we discuss our numerical experiments, this simplification allows D-Flow to meaningfully encode representations of the data with discrete variables, $\s$. The training of D-Flow is also unbiased and performed via likelihood maximization and has fully propagating gradients. This is to be contrasted with other techniques used in the literature to implement generative models with discrete latent variables (especially VAEs), which require biased smoothing of the original model to efficiently propagate gradients through the discrete variables. Notice that the continuous energy model at the base of D-Flow can thus be considered as a smoothing technique for discrete variables, and could be used for the priors of VAEs as well.

    \section{Experiments}
\label{sec:experiments}

\begin{figure*}[t]
\begin{center}
\subfigure[\, Bits per Dim.]{\includegraphics[width=0.32\columnwidth]{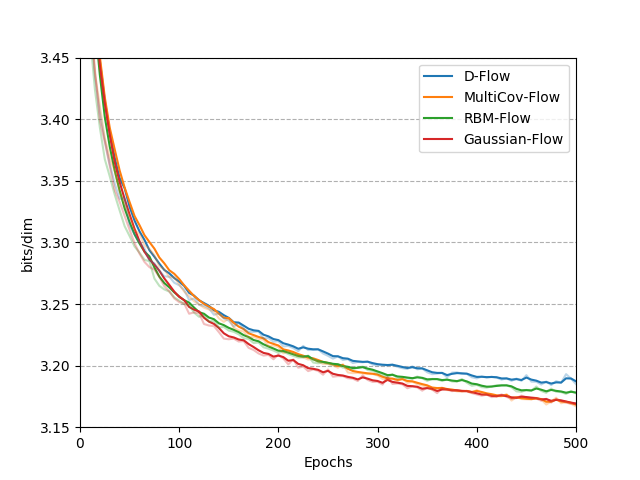}
\label{fig:1a}}
\subfigure[\, FID.]{\includegraphics[width=0.32\columnwidth]{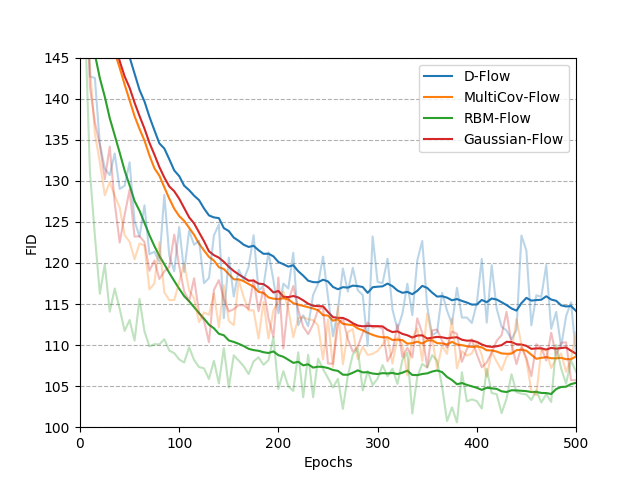}
\label{fig:1b}}
\subfigure[\, IS.]{\includegraphics[width=0.32\columnwidth]{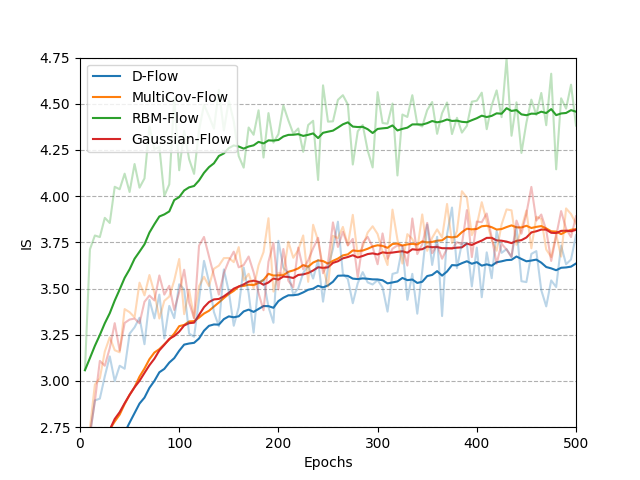}
\label{fig:1c}}
\caption{a) Density estimation performance of each model for the CIFAR10 dataset measured in bits per dimension. For true density estimation of models with EBM base distributions, the partition function was calculated using annealed importance sampling. b) The Frechet inception distance (FID) between the model generated CIFAR10 samples and the test dataset. c) Inception scores (IS) of model generated CIFAR10 samples. Each model has approximately 31.4M parameters and an IF model architecture as described in Appendix~\ref{sec:Appendix A: RBM-IF Architecture}.}
\label{fig:1}
\end{center}
\end{figure*}

\begin{table}[b]
\begin{center}
\begin{tabular}{|lccc|}
\hline
\multicolumn{4}{|c|}{CIFAR10}\\
 \hline
Model & Bits per Dimension  & Inception Score & Frechet Inception Distance \\
\hline
 RBM-Flow (500 epochs)  & $3.18 \pm 0.01$  & $\mathbf{4.4} \pm 0.1$ & $\mathbf{105} \pm 1$ \\
 D-Flow (500 epochs) & $3.19 \pm 0.01$  &$3.2\pm 0.1$ & $114 \pm 1$\\
 MultiCov-Flow  (500 epochs)& $  \mathbf{3.17}\pm 0.01$  &$3.8\pm 0.1$ & $109 \pm 1$\\
 Gaussian Flow  (500 epochs)& $ \mathbf{3.17}\pm 0.01$  &$3.8\pm 0.1$ & $109 \pm 1$\\
\hline
 Flow++  (converged) & $ \mathbf{3.08}$  & - & -\\
 Glow  (converged) & $3.35$  & - & -\\
 Real-NVP  (converged) & $3.49$  & - & -\\
\hline
\end{tabular}
\end{center}
\caption{Image modeling results. While RBM-Flow achieve similar performance to other ablated models in terms of bits/dim, it significantly outperforms them in terms of the Inception Score and the Frechet Inception Distance.}
\label{tab:1}
\end{table}

The goal of our experiments is to understand if an EBM-Flow can improve density modeling of IFs. To do so, we focus on implementing an RBM-Flow, as described in the previous section, and perform a series of ablation experiments in order to determine the role played by the base model in the IF.

The IF used in all models is a custom implementation of Flow++, which is the state-of-the art for density modeling with IFs. The main difference between our implementation and Flow++ is that we do not perform data-dependent initialization, and we do not use a mixed logistic coupling layer in the variational dequantization model. While we have implemented self-attention, we have not used in the results we report due to computational constraints. We describe the flow in more details in Appendix~\ref{sec:Appendix A: RBM-IF Architecture}. 

The models we chose for our comparison are RBM-Flow, D-Flow, a flow with an independent Normal distribution (Gaussian-Flow), and a flow with a multivariate Normal base distribution (MultiCov-Flow). The RBM-Flow $\rightarrow$ D-Flow ablation removes all correlations from the base model but maintains the discrete latent variables. The RBM-Flow $\rightarrow$ MultiCov-Flow ablation keeps the correlations modeled by the full covariance matrix $\Sigma$, but removes the multi modality that can be captured with the latent RBM. Finally, the RBM-Flow $\rightarrow$ Gaussian-Flow ablation removes all modality, covariance and discrete latent variables to be inline with standard approach to density modeling with IFs.
\begin{figure*}[t]
\begin{center}
\subfigure[\, RBM-Flow (500 epochs).]{\includegraphics[width=0.45\columnwidth]{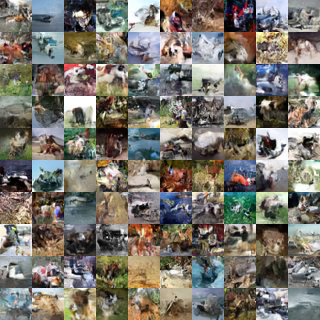}
\label{fig:2a}}
\subfigure[\, IF with Multivariate Gaussian (500 epochs).]{\includegraphics[width=0.45\columnwidth]{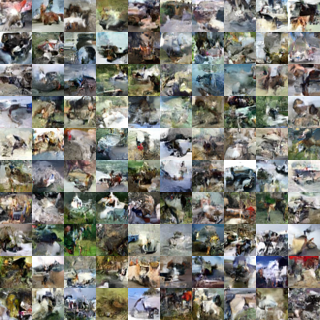}
\label{fig:2b}}\\
\subfigure[\, D-Flow (500 epochs).]{\includegraphics[width=0.45\columnwidth]{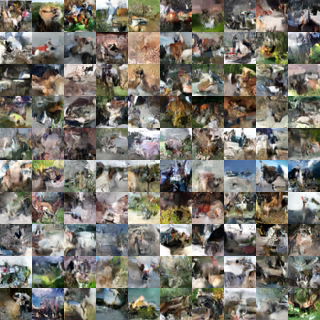}
\label{fig:2c}}
\subfigure[\, IF with independent Gaussian (500 epochs).]{\includegraphics[width=0.45\columnwidth]{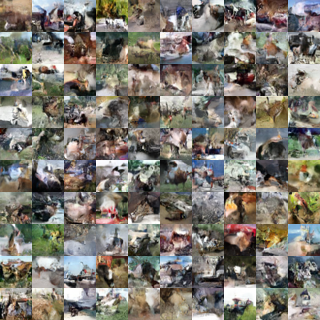}
\label{fig:2d}}
\caption{Qualitative analysis of the generated CIFAR10 samples agree with image quality metrics presented in Fig.~\ref{fig:1}, with their being more structure and color in the images generated using an RBM prior (Fig.~\ref{fig:2a}) compared to other priors, despite all models achieving similar likelihoods.}
\label{fig:2}
\end{center}
\end{figure*}

We have trained all our models on a single NVIDIA V100 GPU by choosing a small batch size of 20 on the CIFAR10 dataset for 500 epochs. All ablation experiments used the same flow architecture, but have differing numbers of filters and components in the mixed logistic coupling layer such that all models have roughly 31.4 million trainable parameters - a number which is similar to that of Flow++ for the CIFAR10 dataset. 

To implement RBM-Flow, we have directly parameterized the weight matrix $\tilde J = J + \Delta \mathbb{1}$, where $\Delta$ is a fixed constant and the $J$ variables are initialized to zero. Using a $\Delta = 2.5$ with L2 normalization of the RBM's weights and biases ensured the positivity of $\tilde J$ in during training of RBM-Flow. This value of $\Delta$ was also used in for the MultiCov-Flow ablation. We have picked $\Delta = 1$ in the case of D-Flow, since we have noticed that with a smaller $\Delta$ D-Flow achieves better likelihood. We found that if $\tilde J$ were to become non positive-definite during training, it typically happened within the first few epochs. To train the RBM of RBM-Flow, we used PCD with 200 block Gibbs updates per gradient update.

\begin{figure*}[t]
\begin{center}
\subfigure[\, RBM-Flow on Fashion MNIST.]{\includegraphics[width=0.45\columnwidth]{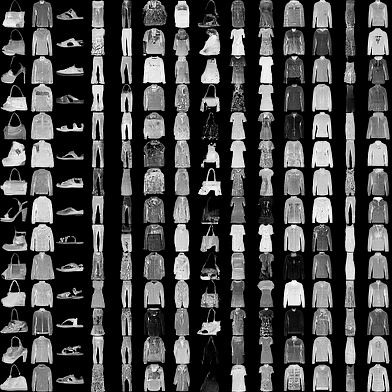}
\label{fig:3a}}
\subfigure[\, D-Flow on Fashion MNIST.]{\includegraphics[width=0.45\columnwidth]{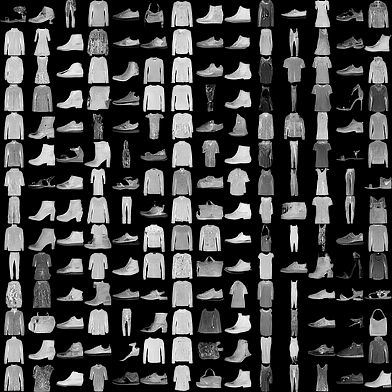}
\label{fig:3b}}\\
\subfigure[\, RBM-Flow on CIFAR10.]{\includegraphics[width=0.45\columnwidth]{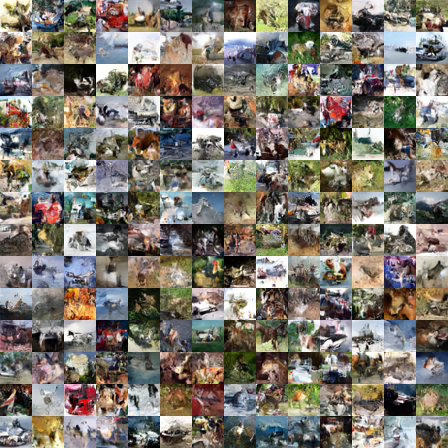}
\label{fig:3c}}
\subfigure[\, D-Flow in CIFAR10.]{\includegraphics[width=0.45\columnwidth]{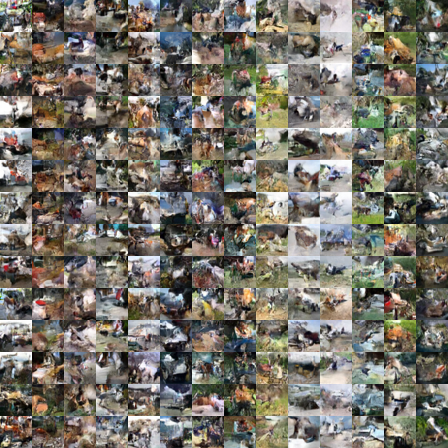}
\label{fig:3d}}
\caption{Generated images for both the Fashion MNIST and CIFAR10 datasets, where each column in every image is a single discrete sample which then conditions the Normal distribution in Eq.~\ref{eq: discrete_smoothing} to give the column of generated images. Discrete samples from both RBM-Flow and D-Flow can encode the global features of a sample in the fashion MNIST dataset. However, RBM-Flow is unable to do this for the CIFAR10 dataset, as each discrete sample is highly correlated to other discrete variables, such that no global features are visually encoded in the discrete sample.}
\label{fig:3}
\end{center}
\end{figure*}

Density modeling results are presented in Fig.~\ref{fig:1a} and Tab.~\ref{tab:1}, and we see that all models achieve similar performance in terms of bits/dim, with Gaussian-Flow and MultiCov-Flow performing lightly better at $3.17$ bits/dim, and RBM-Flow and D-Flow achieving $3.18$ and $3.19$ bits/dim respectively after the same number of epochs. However, these results are sensitive to the value of $\Delta$ used in training, which was partially optimized to the aforementioned values. Attempts to optimize $\Delta$ during training gave rise to instabilities in the model, caused by either diverging $\Delta$ values and/or a non-positive semi-definite covariance matrix.

In Fig.~\ref{fig:2} we present generated samples from all models for comparison and visual inspection. We notice that the samples generated by RBM-Flow appear of higher quality, with more defined global correlation in terms of both color and structure. We quantified this visual perception by computing the Inception Score (IS) and the Frechet Inception Distance (FID). Numerical results are presented in Fig.~\ref{fig:1b}, Fig.~\ref{fig:1c}, and Tab.~\ref{tab:1} where it is shown that that RBM-Flow achieves significantly better IS and FID scores than other ablation models. In particular, Figs.~\ref{fig:1b} and~\ref{fig:1c} show that RBM-Flow is able to achieve higher FID and IS scores much more quickly than other ablation models, a property that can be confirmed also by visually inspecting generated samples. RBM-Flow is the only model to have a base distribution able to capture complex correlations among the base variables $\z$ with a multi modal distribution. The results presented in this section are strong evidence that the smoothed RBM base model is key to improve the overall quality and consistency of the generated samples.

\subsection{Encoding Global Features in Discrete Base Variables with D-Flow}

In certain applications, there might be advantages in developing generative models with discrete latent spaces. Such models could more efficiently learn to encode global features of the original dataset, via unsupervised training, into a discrete representation. This problem has recently been solved in the context of VAEs~\cite{vahdat_dvae_2018, sadeghi_pixelvae_2019, khoshaman_gumbolt_2018, khoshaman_quantum_2018, rolfe_discrete_2017}. To the best of our knowledge, RBM-Flow and D-Flow represent the first implementation of an IF with a discrete latent space. We would like to understand the role played by the discrete variables $\s$, and whether they learn meaningful features.

In Fig.~\ref{fig:3} we show generated samples from models trained on Fashion MNIST and CIFAR10. For each column, we use the same discrete latent space configuration, $\s$, which is then smoothed using Eq.~\ref{eq: discrete_smoothing}. When trained on Fashion MNIST, both RBM-Flow and D-Flow encode global features of the generated samples in the discrete latent configurations. When trained on a more complex dataset such as CIFAR10, D-Flow still encodes global features in the discrete latent space while RBM-Flow does not.  It is not exactly clear why RBM-Flow is less effective than D-Flow at encoding visible features in the discrete latent variables, and it might be worth to investigate this issue more in subsequent work.

    \section{Conclusions}
\label{sec:conclusions}

We have proposed EBM-Flows, i.e. Invertible Flow (IF) models with Energy Based Models (EBM) as trainable base distribution. As we have shown in section~\ref{sec:eflow}, EBM-Flows are themselves EBMs with a scalar energy function that is given by a simple coordinate transformation, given by the IF, of the energy function of the base EBM. EBM-Flows thus represent a class of interesting generative models that can combine the strengths of both IFs and EBMs to produce samples of superior quality.

Using a Restricted Boltzmann machine (RBM) as the EBM, we also introduce an EBM-Flow sub-class called RBM-Flow, which has a Flow++-like architecture and an RBM base model that is continuously smoothed. Implementing RBM-Flow allowed us to perform fast and reliable sampling from the base model, which can be performed via block Gibbs sampling of the underling RBM. This also allowed us to obtain accurate evaluation of the log-likelihood of the model. We have studied the performance of RBM-Flow via a series of ablative experiments, where we keep the number of trainable parameters constant but remove the capability of the base model to represent multi-modal probability distributions. While RBM-Flow performs similarly in terms of log-likelihood to the other ablated models, it generates samples of higher visual quality. We have confirmed this quantitatively by evaluating the Inception Score (IS) and the Frechet Inception Distance (FID) for all models.

Among the ablated models considered, D-Flow is especially interesting. D-Flow is obtained from RBM-Flow by setting to zero all the couplings of the underlying latent RBM. While both RBM-Flow and D-Flow are built with a discrete latent space as part of their base distribution, only D-Flow is able to consistently encode global features of the generated samples into its discrete latent configurations. D-Flow can thus be seen as a genuine implementation of an IF with meaningful discrete latent space representations, generalizing to IFs similar work done with variational autoencoders (VAEs)~\cite{vahdat_dvae_2018, sadeghi_pixelvae_2019, khoshaman_gumbolt_2018, khoshaman_quantum_2018, rolfe_discrete_2017}.

As a future work, it would be interesting to explicitly implement an EBM-Flow model by leveraging recent advances in the implementation of EBMs~\cite{du_implicit_2019, song_generative_2019, gao_flow_2020} and their integration with generative models~\cite{NCP-VAE, VAEBM, gao_flow_2020}. We believe such EBM-Flows could be able to improve the state-of-the-art in generative sampling with standalone EBMs or IFs. Moreover, we notice that the smoothing technique introduced with RBM-Flow and D-Flow could be easily integrated with VAEs. It would be interesting to compare the performance of such models with Gaussian smoothing of RBM and Bernoulli variables to previously introduced VAEs with discrete latent variables.

Another interesting future line of research is the implementation of quantum-assisted training of RBM-Flows with quantum devices. Recent developments with quantum technologies suggest early quantum devices could speed up sampling from Boltzmann machines~\cite{vinci_path_2019}. This idea has motivated the development of quantum/classical hybrid generative models based on VAEs that are amenable to quantum-assisted training. RBM-Flow now gives the possibility to investigate quantum-assisted training of hybrid models based on invertible flows.

    \section*{Acknowledgements}
\label{sec:Acknowledgements}

We are grateful for support from NASA Ames Research Center, the AFRL Information Directorate under grant F4HBKC4162G001, and the Office of the Director of National Intelligence (ODNI) and the Intelligence Advanced Research Projects Activity (IARPA), via IAA 145483. The views and conclusions contained herein are those of the authors and should not be interpreted as necessarily representing the official policies or endorsements, either expressed or implied, of ODNI, IARPA, AFRL, or the U.S. Government. The U.S. Government is authorized to reproduce and distribute reprints for Governmental purpose notwithstanding any copyright annotation thereon.
    
    \medskip
    \printbibliography

    \appendix
    \section{Appendix: RBM-Flow Architecture}
\label{sec:Appendix A: RBM-IF Architecture}

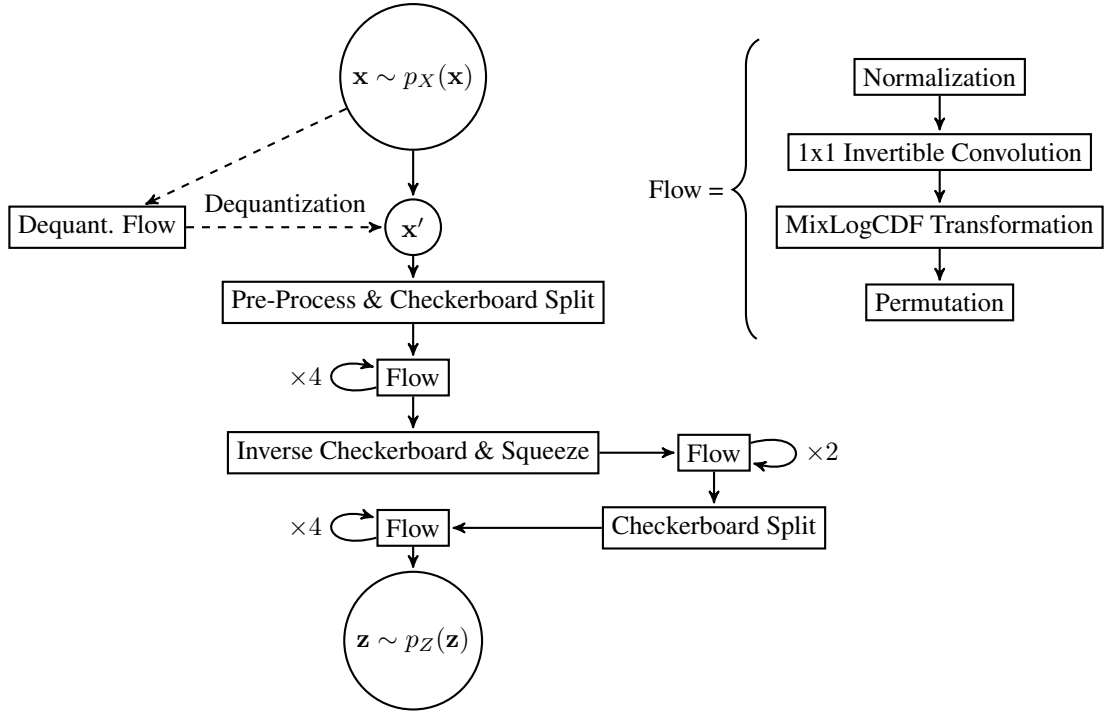
\begin{figure}[h]
	\centering
	\begin{tikzpicture}[-,>=stealth',shorten >=1pt,auto,node distance=1.0cm,
	thick,roundnode/.style={circle,draw},squarenode/.style={rectangle,draw},  el/.style = {inner sep=2pt, align=left, sloped},]

	\node[roundnode] (1) {$\mathbf{x} \sim p_X (\mathbf{x})$};
	\node[roundnode] (2) [below of=1, yshift=-1cm] {$\mathbf{x'}$};
	\node[squarenode] (3) [left of=2,align=center, xshift=-3.2cm] {Dequant. Flow};
	\node[squarenode] (4) [below of=2,align=center] {Pre-Process \& Checkerboard Split};
	\node[squarenode] (6) [below of=4] {Flow};
	\node[squarenode] (7) [below of=6,align=center] {Inverse Checkerboard \& Squeeze};
	\node[squarenode] (9) [right of=7, xshift=3cm] {Flow};
	\node[squarenode] (10) [below of=9,align=center] {Checkerboard Split};
	\node[squarenode] (11) [left of=10, xshift=-3cm] {Flow};
	\node[roundnode] (12) [below of=11, yshift=-0.5cm] {$\mathbf{z} \sim p_Z (\mathbf{z})$};
	
	\node[squarenode] (13) [right of=1, xshift=6cm] {Normalization};
	\node[squarenode] (14) [below of=13] {1x1 Invertible Convolution};
	\node[squarenode] (15) [below of=14] {MixLogCDF Transformation};
	\node[squarenode] (16) [below of=15] {Permutation};
	\draw [decorate,decoration={brace,amplitude=10pt,mirror},yshift=0pt]
    (4.6,0.5) -- (4.6,-3.5) node [black,midway,xshift=-1.6cm] {Flow = };

	\path[->] (1) edge [below] (2);
	\path[dashed,->] (1) edge [left](3);
	\path[->] (2) edge [below]  (4);
	\path[dashed,->] (3) edge [right] node[above] {Dequantization} (2);
	\path[->] (4) edge [below] (6);
	\path[->] (6) edge [loop left] node[left] {$\times 4$} (6);
	\path[->] (6) edge [below] (7);
	\path[->] (7) edge [below] (9);
	\path[->] (9) edge [loop right] node[right] {$\times 2$} (9);
	\path[->] (9) edge [below] (10);
	\path[->] (10) edge [below] (11);
	\path[->] (11) edge [loop left] node[left] {$\times 4$} (11);
	\path[->] (11) edge [below] (12);
	
	\path[->] (13) edge [below] (14);
	\path[->] (14) edge [below] (15);
	\path[->] (15) edge [below] (16);

	\end{tikzpicture}
	\caption{General overview of the framework used in FLOW++ \cite{ho_flow_2019}. The network shown is that used for training the FLOW++ model, but for generation and inference the dequantization flow is removed. Checkerboard split and squeeze processes refer to dimension reshaping processes used in FLOW++ to increase the number of dimensions in the image channel such that the flow can ensure the equal partitioning of inputs.}
	\label{fig:RBM_framework}
\end{figure}

The general framework used for RBM-IF is based off the current state-of-the-art model from FLOW++ \cite{ho_flow_2019}, where the outline of its architecture can be seen in Fig.~\ref{fig:RBM_framework}. The three key features to be included into RBM-IF from FLOW++ were the MixLogCDF transformations (Eq.~\ref{eq: MLCDF}), the variational dequantization model, and the residual gated networks used to support the MixLogCDF transformations. For the most part, each of these features was mirrored in RBM-IF, with exception to minor alterations in the variational dequantization model and residual gated convolutions to reduce training times.

Firstly, the variational dequantization model used in RBM-IF was still composed of 4 conditioned flow processes, each similar to what is seen in the top right of Fig. \ref{fig:RBM_framework}, but with an additional input into the MixLogCDF process to condition the transformation of the noise sample with the data sample. However, in RBM-Flow we replace the MixLogCDF transformation in the dequantization flow with a simpler affine transformation, as it was found not to significantly affect losses and improved training times. Secondly, we omitted the multi-head self attention found in the residual gated convolution process for the purposes of training speed as well.

As detailed in the main text, the RBM prior was the major change in the IF architecture compared to the simpler Gaussian prior used in most IF models. The main overheads incurred by using the RBM prior come from three sources of extra computation. Firstly, matrix inversion and Cholesky decomposition is required in training to define the co-variance matrix of the smoothed RBM prior; secondly, block Gibbs samples are generated in training, and finally, in order to evaluate the true likelihood, the logarithm of the partition function is also calculated, but this is only done once at the end of an epoch. Despite this, all of these sources of additional overhead do not hinder the training process as much as one would expect, such that we still have efficient sampling and density estimation.

It is also worth noting that our reported loss was evaluated using raw test samples, and not with variational dequantization enabled. This meant that we measured against the exact likelihood instead of a variational bound introduced by the dequantization, so our results cannot be directly compared to FLOW++ as we do not use importance sampling.

Finally, we compensate for the additional parameters incurred by using energy based priors by changing the number of parameters in the IF model. This is done by changing the number of convolutional filters and the number of components in the MixLogCDF transformation, which is detailed in Tab.~\ref{tab:prior-parameters}. Other model hyper-parameters include a constant learning rate of $10^{-4}$, batch size of $20$, and a dropout rate of $0.2$. 

\begin{table}[h]
    \centering
    \begin{tabular}{c c c c}
        \toprule
         & Gaussian & RBM & Bernoulli \\
         \midrule
         No. Convolutional Filters & 100 & 96 & 100 \\
         MixLogCDF Components & 33 & 32 & 33 \\
         No. Gated Convolutions & 10 & 10 & 10 \\
         No. Trainable Parameters & 31,445,192 & 31,420,584 & 31,448,264 \\
         \bottomrule \\
    \end{tabular}
    \caption{Number of model parameters used for each prior in testing with CIFAR10, and their respective hyper-parameters used. All other aspects of each model are identical.}
    \label{tab:prior-parameters}
\end{table}

The code for the RBM-IF model was created within TensorFlow(TF) v2.1.0 using TensorFlow Probability (TFP) v0.9.0 bijectors contained within custom Keras layers. A custom training loop is also used in order to implement conditional transformations for the variational dequantization model, as well as for the passing of training arguments, which not all TFP bijectors support. 
\typeout{get arXiv to do 4 passes: Label(s) may have changed. Rerun}
\end{document}